\documentclass[twoside,11pt]{article}

\usepackage{amsmath}
\usepackage{amsfonts}
\usepackage{amssymb}
\usepackage{mathtools}
\usepackage{graphicx}
\usepackage{subfigure}
\usepackage{color}
\usepackage[numbers]{natbib}
\usepackage{snapshot}

\allowdisplaybreaks[1]

\renewcommand{\vec}[1]{\boldsymbol{\mathrm{#1}}}
\newcommand{\rank}{\operatorname{rank}}
\newcommand{\diag}{\operatorname{diag}}
\newcommand{\T}{\mathrm{T}}
\renewcommand{\d}{\mathrm{d}}

\newcommand{\partialfrac}[3][]{\frac{\partial^{#1} #2}{\partial #3^{#1}}}
\newcommand{\ie}{i.e.\ }
\newcommand{\eg}{e.g.\ }

\renewcommand{\cite}{\citet}

\title{A Characterization of the Combined Effects of Overlap and Imbalance on
  the SVM Classifier}
\date{}
\author{}

\begin{document}



\maketitle

\begin{center}
\textbf{\large{Misha Denil}}
and
\textbf{\large{Thomas Trappenberg}}\\
Dalhousie University\\
$^{1}$\texttt{denil@cs.dal.ca}\\
$^{2}$\texttt{tt@cs.dal.ca}\\
\end{center}

\vspace{0.5cm}


\begin{abstract}
  In this paper we demonstrate that two common problems in Machine
  Learning---imbalanced and overlapping data distributions---do not have
  independent effects on the performance of SVM classifiers.  This result is
  notable since it shows that a model of either of these factors must account
  for the presence of the other.  Our study of the relationship between these
  problems has lead to the discovery of a previously unreported form of
  ``covert'' overfitting which is resilient to commonly used empirical
  regularization techniques.  We demonstrate the existance of this covert
  phenomenon through several methods based around the parametric regularization
  of trained SVMs.  Our findings in this area suggest a possible approach to
  quantifying overlap in real world data sets.
\end{abstract}


\section{Introduction}

A data set is imbalanced when its elements are not evenly divided between the
classes. In practical applications it is not uncommon to see very high
imbalance, where upwards of 90\% of the available training data belong to only
one class. Overlap is another common problem, which occurs when there are
regions of the data space where the posterior class distributions are near
equal, even when the priors are known with certainty.  In these cases it is
difficult to make a principled decision on how to divide the volume of these
regions between the classes.

Although the overlap and imbalance problems have been studied previously, (see
\cite{Weijters1997}; \cite{Japkowicz2002}; \cite{Akbani2004}; \cite{Monard};
\cite{Yaohua2007} for some representative works in this area), work on each
problem has happened largely in isolation.  Some authors (\eg \cite{Auda1997};
\cite{Visa}; \cite{Prati2004}; and \cite{Batista2005}) have performed
experiments in the presence of both factors; however, the nature of their
interaction is still not well understood.  Our finding that their effects are
not independent is an important step towards a characterization of how these
factors affect classifier performance.

We propose that the behaviour observed in the combined case can be explained by
phenomenon we call ``covert'' overfitting.  Covert overfitting is similar in
principle to regular overfitting, but the ambiguities which lead to overfitting
are present in the generative distributions of the classes, rather than just in
the training set.  This complication ensures that standard empirical
regularization techniques, such as cross validation, or using a separate
validation set for testing, are not able to detect this phenomenon.  We explore
this problem in detail, and offer several demonstrations of its occurrence, in
the later sections of this paper.

In the first part of this paper we explore how the Support Vector Machine (SVM)
classifier performs when faced with overlapping and imbalanced data sets.  In
contrast to previous work in this area, we directly address the question of how
the relationship between these factors affects classifier performance.  A key
result of this work is that the effects from these factors are not independent.
We show that, although neither factor acting alone has an unexpectedly strong
effect, the presence of overlap and imbalance together causes performance
degradation which is more severe than we are lead to expect by considering them
independently.  This is an extension of our previous work on the overlap and
imbalance problems in \cite{Denil2010b}, but goes beyond it by offering an
explanation and application of the combined effects.  We also demonstrate how
different signatures of these effects might be used as tools to measure overlap
in real world data.


\section{Data and Experimental Setup}
\label{sec:data-sets}

\begin{figure}[htb]
  \centering
  \includegraphics[width=0.95\linewidth]
  {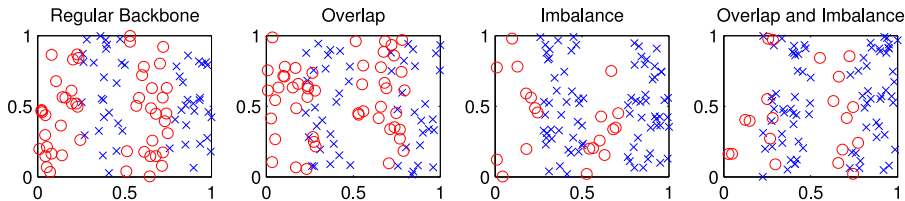}
  \caption{Sample backbone models in two dimensions.}
  \label{fig:sample-backbone}
\end{figure}

We build our analysis around a series of synthetic data sets in the from of two
dimensional ``backbone'' models.  To generate a data set we sample points form
the region $[0,1]\times[0,1]$.  The range along one dimension is divided into
four regions with alternating class membership, (two regions for each class),
while the two classes are indistinguishable in the other dimension (see
Figure~\ref{fig:sample-backbone}).  These domains make a good candidate for
study since they are relatively simple, both to visualize and to understand, yet
the optimal decision boundary is sufficiently non-linear to cause interesting
effects to emerge.  The main problems we discuss in this paper often do not
appear in very simple domains; we have chosen our models to be sufficiently
complex to demonstrate the issues at hand.

Throughout this paper it will be necessary for us to have a parameterization of
the overlap and imbalance levels present in a particular data set.  This will
allow us to study classifier performance with respect to these parameters and to
formulate a model of how they affect performance.

We parameterize the overlap level with $\mu \in [0,1]$ such that when $\mu=0$
the two classes are completely separable and when $\mu=1$ both classes are
distributed uniformly across the entire domain.  Intermediate values of $\mu$
indicate overlap along the region boundaries.

The imbalance level, which we denote $\alpha \in [0.5, 1]$, is measured as the
proportion of the data set belonging to the majority class.\footnote{We only
  consider $\alpha\le 0.95$ in our experiments since, by this parameterization,
  $\alpha=1$ corresponds to a data set with only one class present.}  When there
is imbalance, we always take the second class as the majority class; however,
since the class distributions are symmetric, in these models the distinction
between ``first'' and ``second'' is somewhat arbitrary, hence our decision to
consider only the degree of imbalance and ignore which particular class is
present in the majority.

Using this scheme, we generate a series of data sets for each collection of
experiments by varying one, or both, of the available parameters.  Unless
otherwise indicated, all our experiments are repeated using training sets of
several different sizes varying (logarithmically) between 25 and 6400 examples
(although in the interest of saving space we report only a subset of these
results).  Testing is done using newly generated data sets of the appropriate
imbalance level, overlap level and size.

We assess classifier performance using the F$_1$-score of the classifier trained
on each data set, where the minority class is taken to be positive. The
F$_1$-score is the harmonic mean of the precision and recall of a classifier and
is a commonly used scalar measurement of performance. Our choice of positive
class reflects the state of affairs present in many real world problems where it
is difficult to obtain samples from the class of interest. The F$_1$-score is
one of the family of F$_\beta$-scores and treats precision and recall as equally
important.

Our experiments here focus on the SVM classifier with an RBF kernel.  In all
cases parameter selection for the SVM was carried out using the simulated
annealing procedure described in \cite{Boardman2006} to select
optimal values for $C$ and $\gamma$.


\section{Overlap and Imbalance in Isolation}

In this section we look at how overlap and imbalance in isolation affect
classifier performance.  The purpose of this section is to provide some baseline
results which will inform our analysis of the combined effects in
Section~\ref{sec:combined}.


\subsection{Imbalance}
\label{sec:imbalance}

This section shows a series of experiments using varying levels of imbalance.
We confirm previous results from \cite{Japkowicz2002}, which indicate that
imbalance in isolation is not sufficient to degrade performance.  This suggests
that poor performance on imbalanced data sets is caused by other factors such as
small disjuncts.  (For a discussion of why the imbalance problem is best viewed
as an instance of the small disjunct problem see \cite{Japkowicz2002},
\cite{Japkowicz2003}, and \cite{Jo2004}).

\begin{figure}[tbh]
  \centering
  \subfigure[]{
    \includegraphics[width=0.3\linewidth]
    {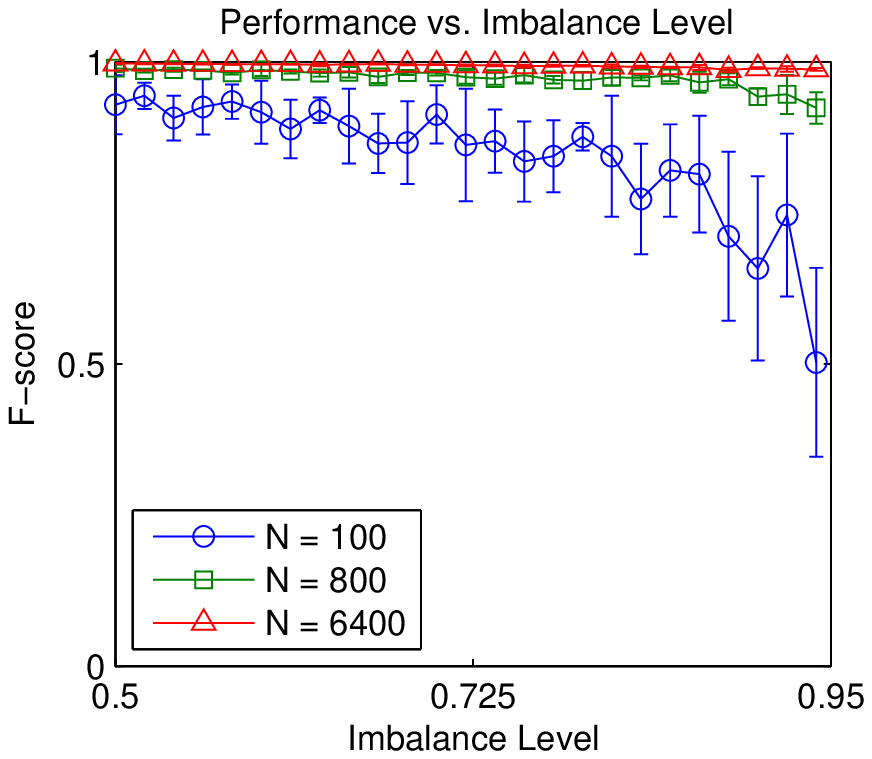}
    \label{fig:perf-v-imbalance}
  }
  \subfigure[]{
    \includegraphics[width=0.3\linewidth]
    {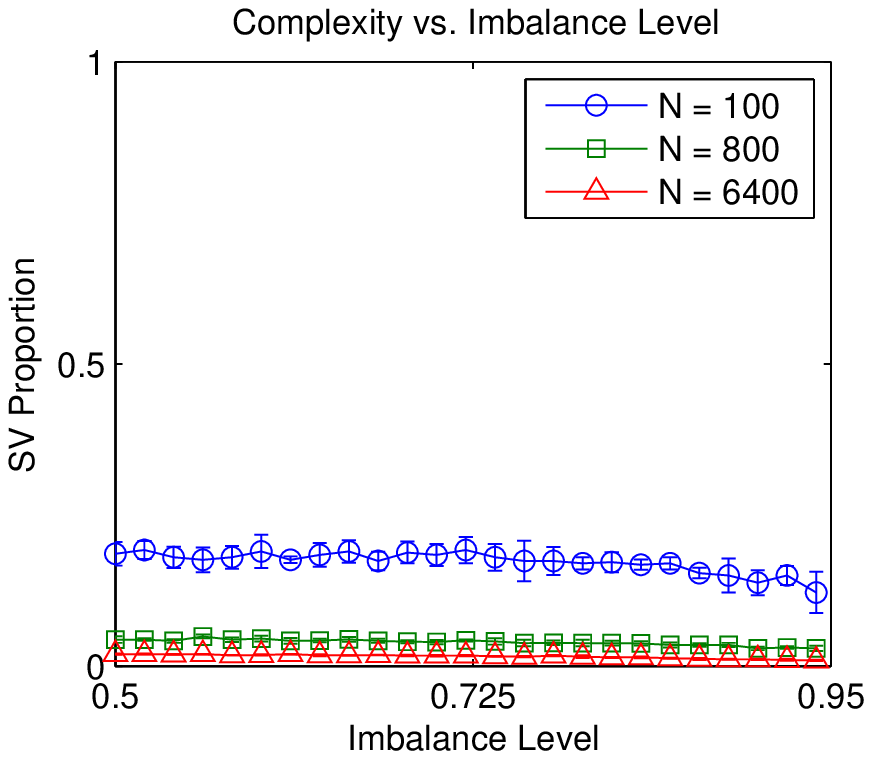}
    \label{fig:complexity-v-imblanace}
  }
  \subfigure[]{
    \includegraphics[width=0.3\linewidth]
    {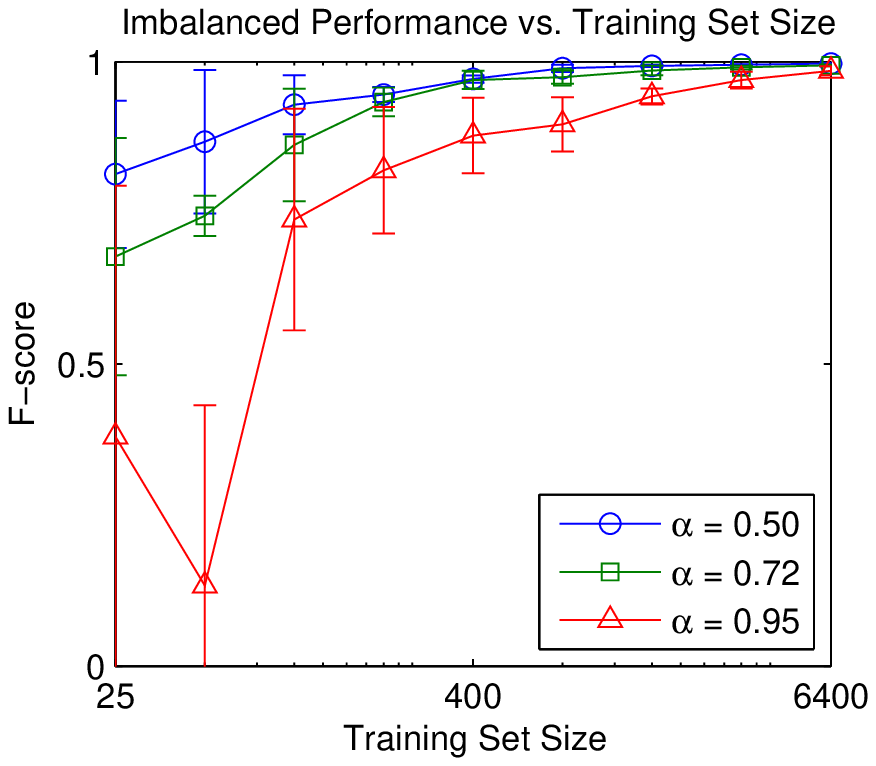}
    \label{fig:imbalance-perf-v-tss}
  }
  \caption{Imbalance in isolation. (a) Shows the SVM performance over a range of
    imbalance levels.  (b) Shows the solution complexity over the same range.
    (c) Shows how performance varies with training set size at various levels of
    imbalance.  In each case $N$ is the number of training data used.  Error
    bars show one standard deviation about the mean over 10 trials. $N$ is the
    number data in the training and test sets.}
  \label{fig:imbalance-isolation}
\end{figure}

Performance results from our experiments are shown in
Figure~\ref{fig:perf-v-imbalance}.  When the training set size is large we
observe that the imbalance level has very little effect on the classifier
performance.  Performance is only affected when either the imbalance level is
very high (and then only slightly), or when there are very few training data.
This is exactly what we expect from the existence of small disjuncts in these
domains.  The influence that the training set size has on performance can be
seen explicitly in Figure~\ref{fig:imbalance-perf-v-tss}.

In addition to the F$_1$-scores, we also recorded the number of support vectors
from each run as a measure of the complexity of the trained
models. Figure~\ref{fig:complexity-v-imblanace} shows the proportion of the
training set retained as support vectors and that the imbalance level has no
visible adverse effect on the complexity of the SVM solution.  In fact, there is
a slight drop in complexity when the imbalance level is very high; however, at
high levels of imbalance there are very few training data available to support
the minority side of the boundary.  This interpretation is supported by the fact
that as the training set size increases the overall proportion that is retained
drops, and the complexity reduction at high imbalance levels becomes less
apparent.

The major conclusion that we can draw here is that imbalance in isolation has no
adverse affect on the SVM classifier, provided that the training set is
sufficiently large.  The reduced performance we see when the training set is
small can be attributed to the fact that there are not sufficiently many
minority examples to infer the class distribution.  This is confirmed by the
fact that with a large training set the performance is excellent, even on highly
imbalanced domains.


\subsection{Overlap}
\label{sec:overlap}

In contrast to the imbalance problem, the effects of overlap are not well
characterized in the literature (although previous work on the problem can be
found in \cite{Visa}; \cite{Prati2004}; and \cite{Yaohua2007}).  We use this
section to demonstrate that overlapping classes cause the SVM to learn decision
boundaries which lack parsimony.

\begin{figure}[tbh]
  \centering
  \subfigure[]{
    \includegraphics[width=0.3\linewidth]
    {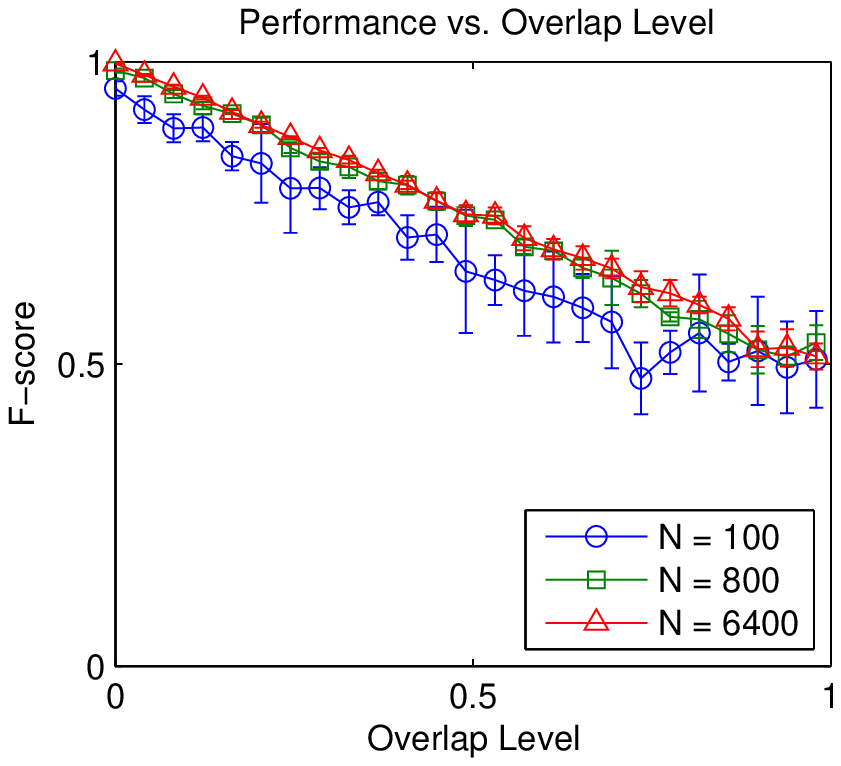}
    \label{fig:perf-v-overlap}
  }
  \subfigure[]{
    \includegraphics[width=0.3\linewidth]
    {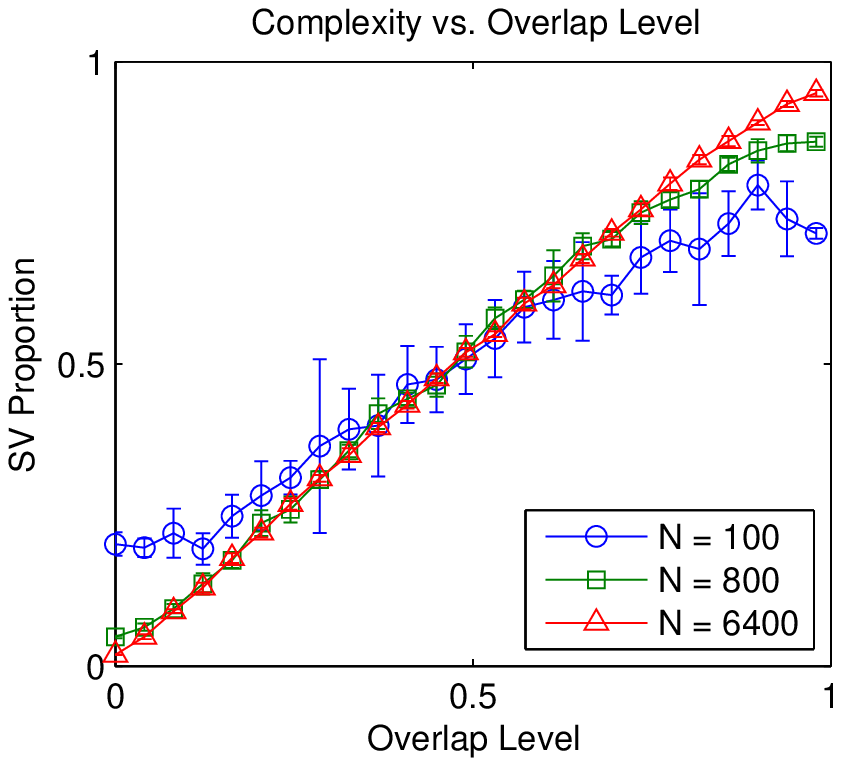}
    \label{fig:complexity-v-overlap}
  }
  \subfigure[]{
    \includegraphics[width=0.3\linewidth]
    {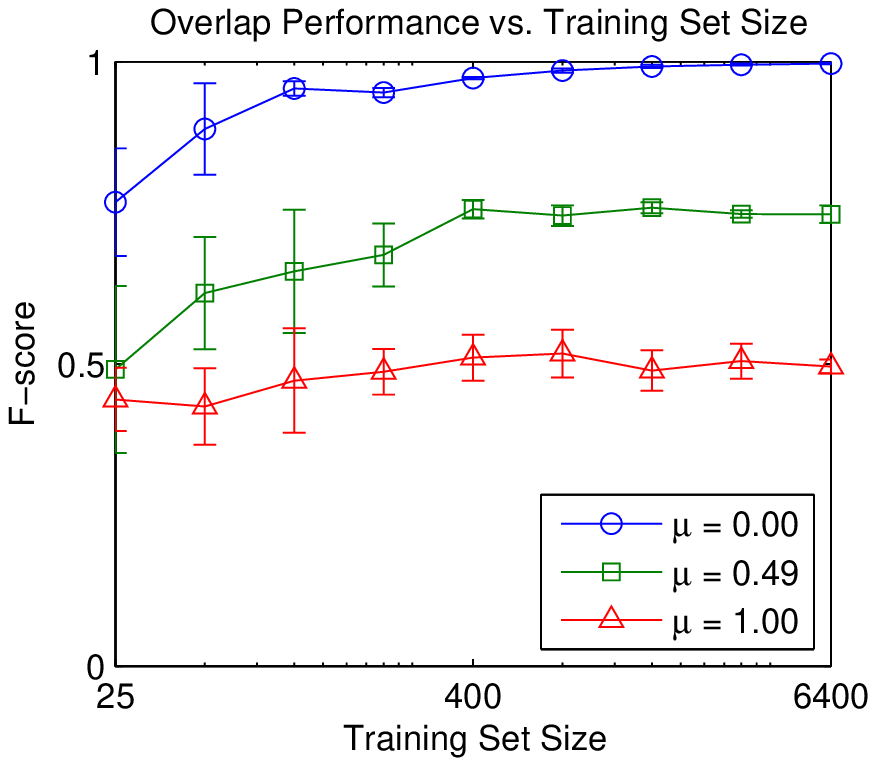}
    \label{fig:overlap-perf-v-tss}
  }
  \caption{Overlap in isolation.  (a) Shows the SVM performance over a range of
    overlap levels.  (b) Shows the solution complexity over the same range.  (c)
    Shows how performance varies with training set size at various levels of
    overlap.  In each case $N$ is the number of training data used.  Error bars
    show one standard deviation about the mean over 10 trials.  $N$ is the
    number of data in the training and test sets. }
  \label{fig:overlap-isolation}
\end{figure}

Figure~\ref{fig:perf-v-overlap} shows performance results with respect to
overlap level for a selection of training set sizes, with the explicit
relationship between training set size and performance appearing in
Figure~\ref{fig:overlap-perf-v-tss}.  The experiments which produced these data
follow the same procedure as those from Section~\ref{sec:imbalance}, but here we
vary the overlap level instead of the imbalance.  As in the case of imbalance,
we see that very small training sets tend to cause degraded performance;
however, in this case the effect is much weaker and becomes less pronounced as
the overlap level is increased (see Figure~\ref{fig:overlap-perf-v-tss}).  This
indicates that, unlike the case of imbalance, when the overlap level is high, it
is unlikely that collecting more training data will produce a more accurate
classifier.

In Figure~\ref{fig:perf-v-overlap} we see that performance of the SVM classifier
in the presence of overlap shows a linear drop as the overlap level is
increased, with the linearity becoming more pronounced with larger training
sets.  An important observation here is that this is precisely what we expect
from an optimal classifier on these domains.  When we introduce overlap into
these (balanced) data sets we create ambiguous regions in the data space where
the generative distributions for both classes are near equal.  This means that
even a classifier with perfect knowledge of the generative distributions will
infer near-equal posterior probabilities in these regions, meaning that we
cannot predict the class label better than chance.

It is more interesting here to examine the complexity of the SVM solutions,
which we again measure using the proportion of the training set retained as
support vectors (shown in Figure~\ref{fig:complexity-v-overlap}).  The response
here again appears linear, but in this case a linear response is somewhat
alarming.  The proportion of the training set retained as support vectors rises
linearly as a function of the overlap level, and this effect is visible across a
wide range of training set sizes.  This indicates that increasing the size of
the training set, which was a boon in the case of imbalance, actually causes the
SVM solution to increase in complexity.

When overlap is present in isolation, the SVM classifier is able to achieve
approximately optimal performance across a wide range of different training set
sizes; however, despite the near optimal performance, as the overlap level is
increased the complexity of the model rises sharply, both as a function of the
overlap level and also as a function of the training set size.  This is
counter-intuitive, as we generally expect that increasing the amount of training
data should lead to ``better'' models.  Due to how we introduce overlap into our
distributions the complexity of the optimal solution is independent of the
overlap level.  


\section{Combined Overlap and Imbalance}
\label{sec:combined}

We now turn our attention to the behavior of the SVM in the presence of both
overlap and imbalance simultaneously.  We are interested in determining if it is
possible to separate the contributions from each factor.  If this is possible
then we can assign blame for different portions of the performance degradation
to each factor; however, if the effects of the two factors interact, this
assignment of blame becomes much more complicated and less useful.

If the effects are independent (\ie they do not interact) then the overlap and
imbalance problems can reasonably be studied in isolation; however, if they are
not independent it is important to understand the relationship between them,
which can only come from studying them together.  We will show that this is in
fact the case, and our study of the combined effects gives rise to the discovery
of a previously unreported phenomenon which we call covert overfitting.


\subsection{Test for Independence}
\label{sec:test-for-indep}

We first outline a method to test the hypothesis that overlap and imbalance have
independent effects on classifier performance.  Let us continue to use $\mu$ as
a measure of overlap and $\alpha$ as a measure of imbalance.  The hypothesis can
be expressed mathematically as the assumption that the performance surface with
respect to $\mu$ and $\alpha$ obeys the relation
\begin{align*}
  \d P(\mu,\alpha) = f'(\mu)\,\d\mu + g'(\alpha)\,\d\alpha
  \enspace ,
\end{align*}
where $f'$ and $g'$ are unknown functions.  That is, we expect the total
derivative of performance to be separable into the components contributed by
each of $\mu$ and $\alpha$.  This hypothesis of independence leads us to expect
that we can consider the partial derivatives as functions of a single variable,
\ie
\begin{align*}
  \partialfrac{}{\mu} P &= f'(\mu) \enspace, \\
  \partialfrac{}{\alpha} P &= g'(\alpha) \enspace. 
\end{align*}

The functions $f'$ and $g'$ may not have simple or obvious functional forms,
meaning that we cannot compute their values analytically; however, if $f'$ and
$g'$ are known we can find a predicted value for $P(\alpha,\mu)$, up to an
additive constant, by evaluating
\begin{align}
  P(\mu,\alpha) = \int f'(\mu)\,\d\mu + \int g'(\alpha)\,\d\alpha + C \enspace
  . \label{eq:P}
\end{align}

Specific values for $P(\mu,\alpha)$ can be computed numerically by training a
classifier on a data set with the appropriate level of overlap and imbalance.
Since we expect the partial derivatives of $P(\mu,\alpha)$ to be independent, we
can compute values for $f'$ by evaluating $P(\mu,\alpha)$ for several values of
$\mu$ while holding $\alpha$ constant and taking a numerical derivative.  Values
for $g'$ can be computed in a similar manner by holding $\mu$ constant and
varying $\alpha$.  These values can then be combined into predicted values for
$P(\mu,\alpha)$ using (\ref{eq:P}).  Comparing the predicted values for
$P(\mu,\alpha)$ to the observed values will allow us to determine if our
hypothesis of independence is sound.

\begin{figure}[tbh]
  \centering
  \includegraphics[width=0.5\linewidth]{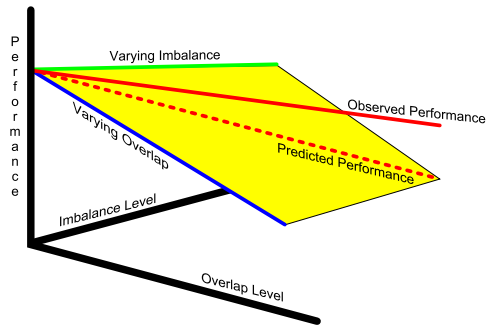}
  \caption{Diagram of the proposed independence test.}
  \label{fig:perf-surf}
\end{figure}

The procedure for applying this model is illustrated in
Figure~\ref{fig:perf-surf}, which shows a performance surface parameterized by
the overlap and imbalance levels of the training set.  First, we take
measurements of this surface along the indicated axis-aligned sections.  This
corresponds to measuring the effects of each factor in isolation, the results of
which where shown in previous sections.  These data, combined with the model of
independence we have described here, allow us to make predictions for the
combined case (the dashed line in Figure~\ref{fig:perf-surf}).  Comparing these
predicted values to the performance of actual classifiers trained on data sets
with the corresponding levels of overlap and imbalance enables us to assess the
correctness of the model.  What we are looking for is a discrepancy between the
model's predictions and our observations (shown in the figure as the difference
between the solid and dashed lines).  If the predictions do not match well with
our observations we can reject the model and conclude that there must be an
interaction between the effects of overlap and imbalance on SVM performance.


\subsection{Results}

\begin{figure}[tbh]
  \centering
  \subfigure[] {
    \includegraphics[width=0.3\linewidth]
    {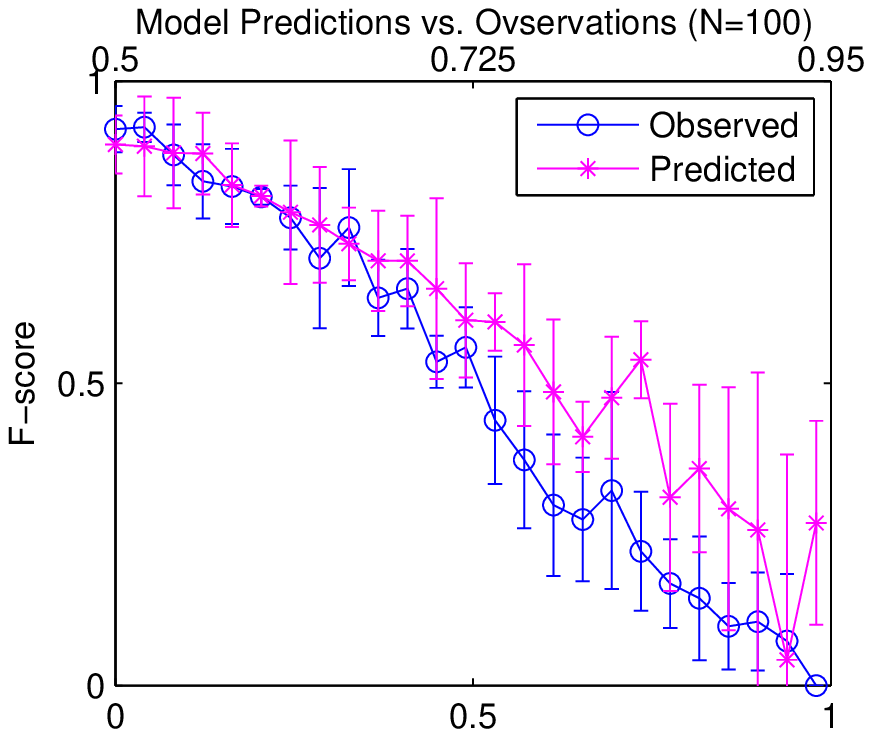}
  }
  \subfigure[] {
    \includegraphics[width=0.3\linewidth]
    {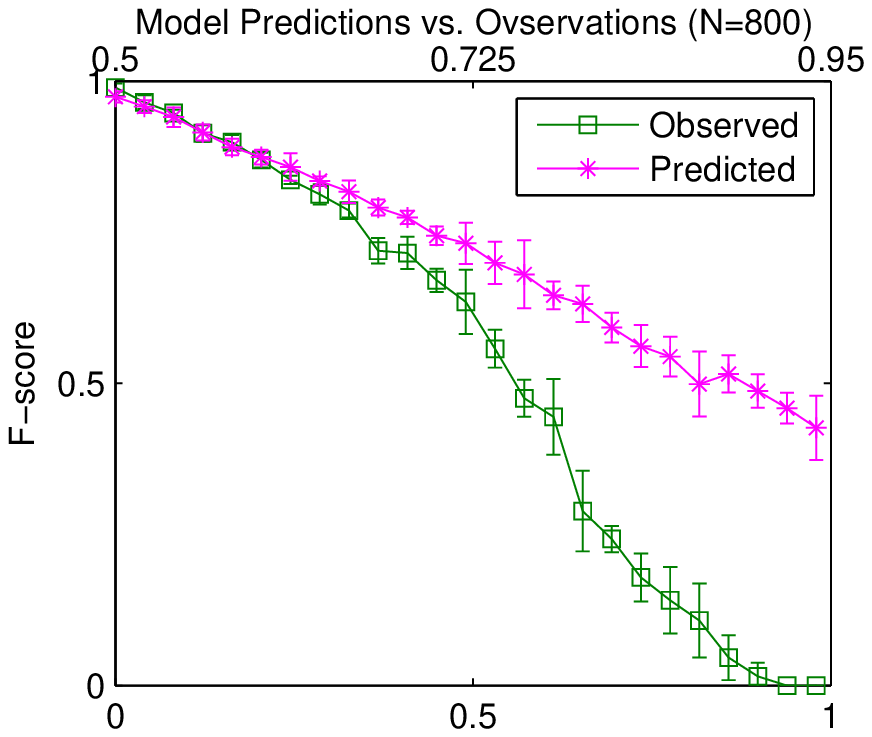}
  }
  \subfigure[] {
    \includegraphics[width=0.3\linewidth]
    {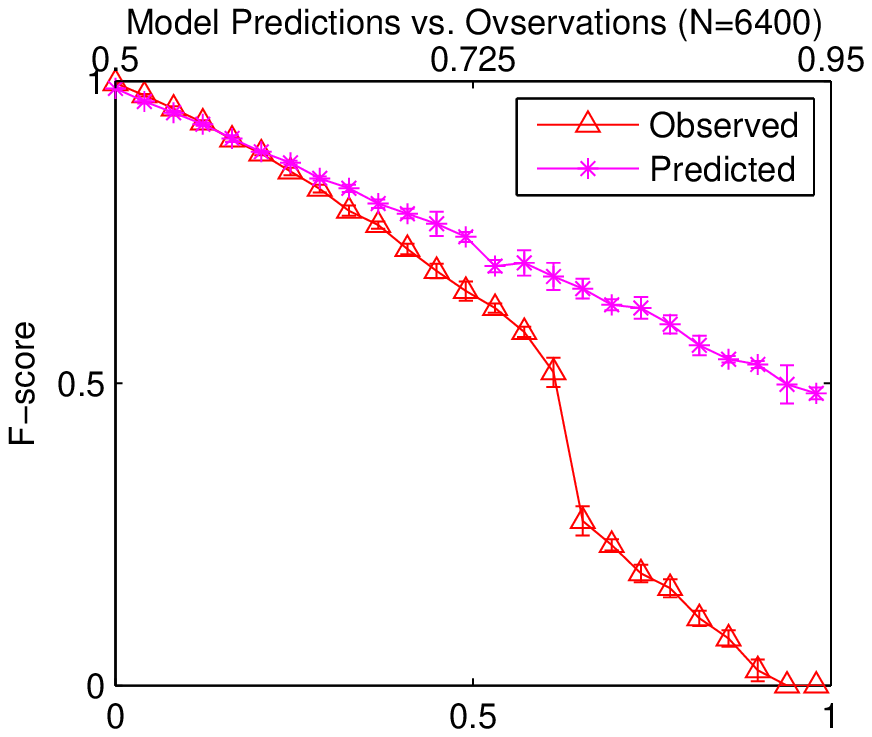}
  }
  \caption{Comparing model predictions to observations in the combined case.  In
    these figures the lower x-axis shows the degree of overlap and the upper
    x-axis shows the degree of imbalance.  $N$ is the number of data in the
    training and test sets.}
  \label{fig:combined-model}
\end{figure}

Comparisons between our model predictions and the observed performance on
domains with combined overlap and imbalance are shown in
Figure~\ref{fig:combined-model}.  These results clearly show that when the
training set size is large, the performance predicted by assuming that overlap
and imbalance are independent is very different than what is observed.  On the
other hand, when the training set is small the predictions are quite accurate,
showing only a small (but still significant) deviation from the observed
results.

In addition to showing performance which falls short of our model's predictions,
we see a sudden breaking point in performance beyond a certain level of combined
overlap and imbalance.  This effect is most pronounced when the training set is
large, becoming less noticeable with fewer training data and disappearing
entirely when the training set size is very small.  This drop occurs
consistently at approximately $\mu=0.6$ and $\alpha=0.78$ with very little
variation across different training set sizes.  In \cite{Denil2010b} we showed
that the differences are statistically significant and that the drop is
correlated with the peak complexity of these models.

\begin{figure}[tbh]
  \centering
  \subfigure[] {
    \includegraphics[width=0.3\linewidth]
    {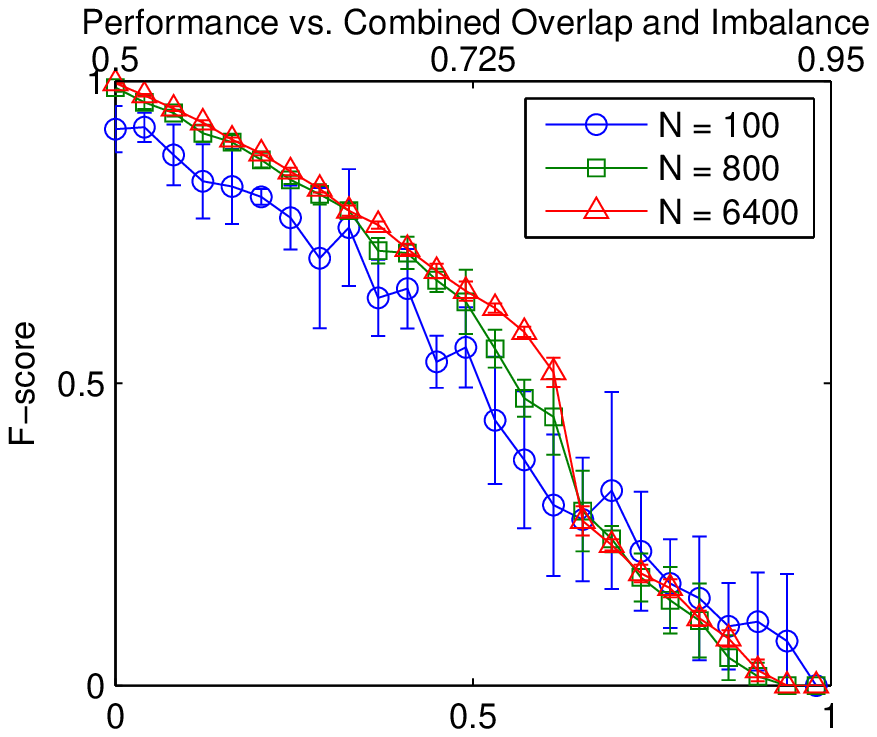}
  }
  \subfigure[] {
    \includegraphics[width=0.3\linewidth]
    {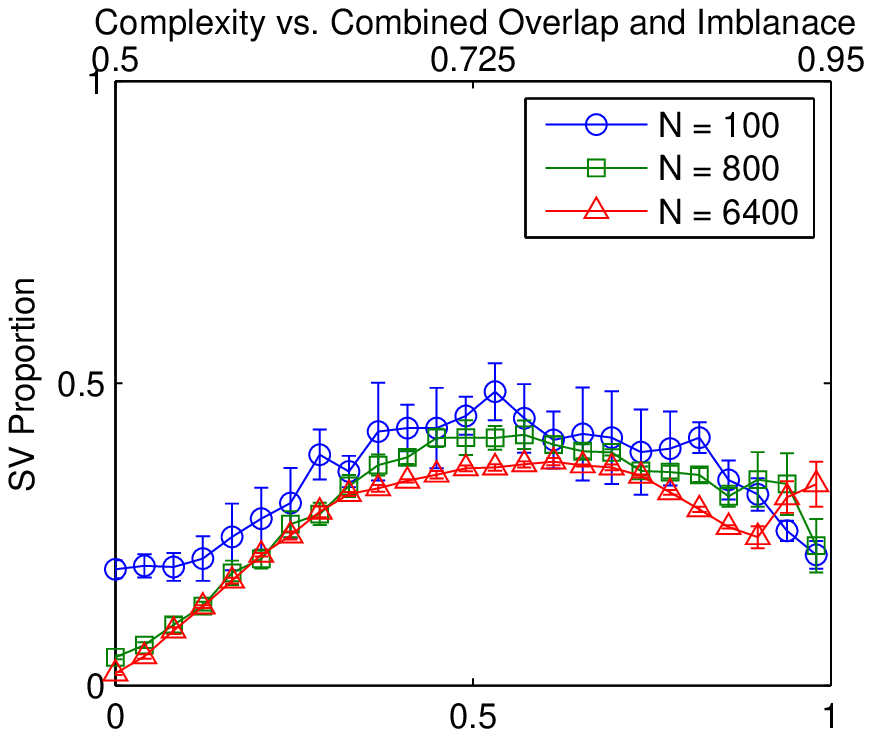}
  }
  \subfigure[] {
    \includegraphics[width=0.3\linewidth]
    {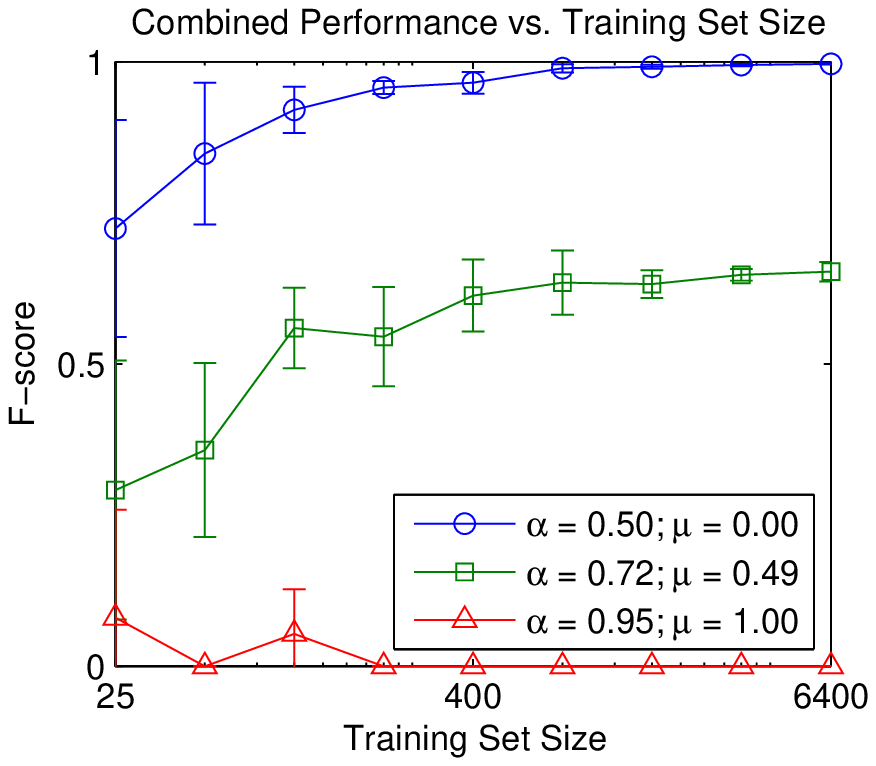}
  }
  \caption{Combined overlap and imbalance.  In these figures the lower x-axis
    shows the degree of overlap and the upper x-axis shows the degree of
    imbalance.  $N$ is the number of data in the training and test sets.}
  \label{fig:combined}
\end{figure}

Figure~\ref{fig:combined} shows the performance and complexity we observed in
the combined case across several training set sizes.  The data are presented
here in the same format as Figures~\ref{fig:imbalance-isolation}
and~\ref{fig:overlap-isolation} for ease of comparison.  These figures emphasize
the breaking point in performance we see with combined overlap and imbalance.
Crucially, we see that the performance beyond this breaking point is unchanged
across the range of training set sizes we tested; however, more data can
significantly improve the pre-breaking-point performance.

The model from Section~\ref{sec:test-for-indep} relies only on the independence
of the imbalance and overlap problems in order to make predictions for
performance in the combined case.  Since we have shown that the model
predictions are very poor, it is reasonable to conclude that the underlying
assumption is incorrect; specifically, we claim that our results demonstrate
that there is an interdependence between the effects of overlap and imbalance.
The later sections of paper are devoted to characterizing this interdependence.


\section{Covert Overfitting}
\label{sec:covert-overfitting}

In this section we propose an explanation for the performance and complexity
behaviours we observe in the presence of overlap and imbalance.  So far we have
seen that:
\begin{itemize}
\item Imbalance, in isolation, is not a significant problem for SVMs.  When
  there are sufficiently many training data available the SVM forms simple
  models (as expected, given the simplicity of our domains) which show excellent
  performance, even when the degree of imbalance is very high.
\item Overlap, in isolation, causes SVMs to build very complex models which
  exhibit performance comparable to an optimal classifier.  Although performance
  drops as the overlap level is increased, it is still optimal since the
  presence of overlap creates ambiguous regions where even an optimal classifier
  cannot predict the class label better than chance.  However, the complexity of
  these models is extremely high, especially considering that the complexity
  required to achieve this performance is no different from the separable case.
\item When both factors are present in tandem not only does the SVM build overly
  complex models, as in the case of overlap in-isolation, but the performance on
  these domains is also significantly reduced.
\end{itemize}
Since the underlying reasons for the behaviour in the case of imbalance in
isolation is fairly well understood (see the beginning of
Section~\ref{sec:imbalance} for references) we will focus on the remaining two
cases here.

We hypothesize that the observed behaviour is a result of a phenomenon we call
\emph{covert overfitting}.  Covert overfitting is similar to ordinary
overfitting, in that it is a result of mistaking aberrations in the training
data for characteristics of the generative class distributions.  The key
difference is that covert overfitting occurs in the ambiguous regions caused by
overlap.

\begin{figure}[tbhp]
  \centering
  \hspace{\stretch{1.3}}
  \includegraphics[width=0.35\linewidth]
  {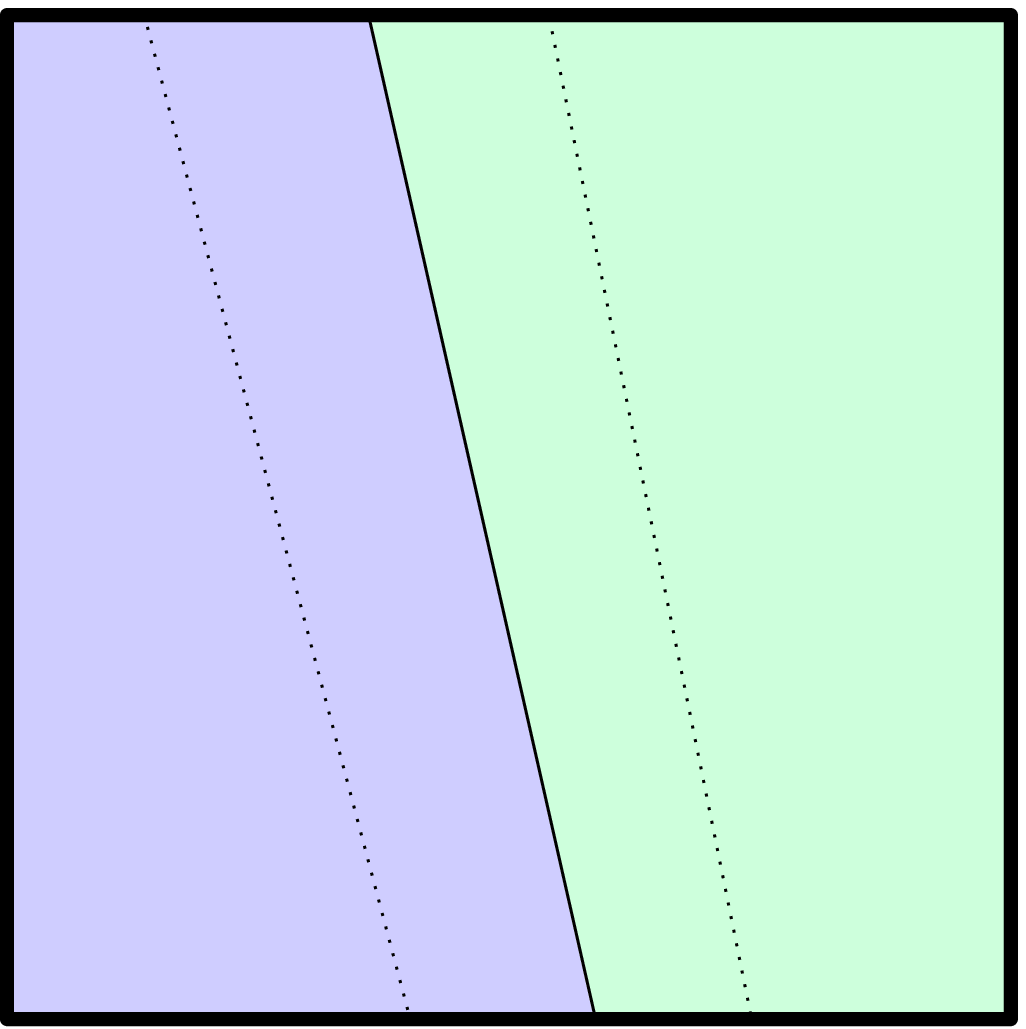}
  \hspace{\stretch{1}}
  \includegraphics[width=0.35\linewidth]
  {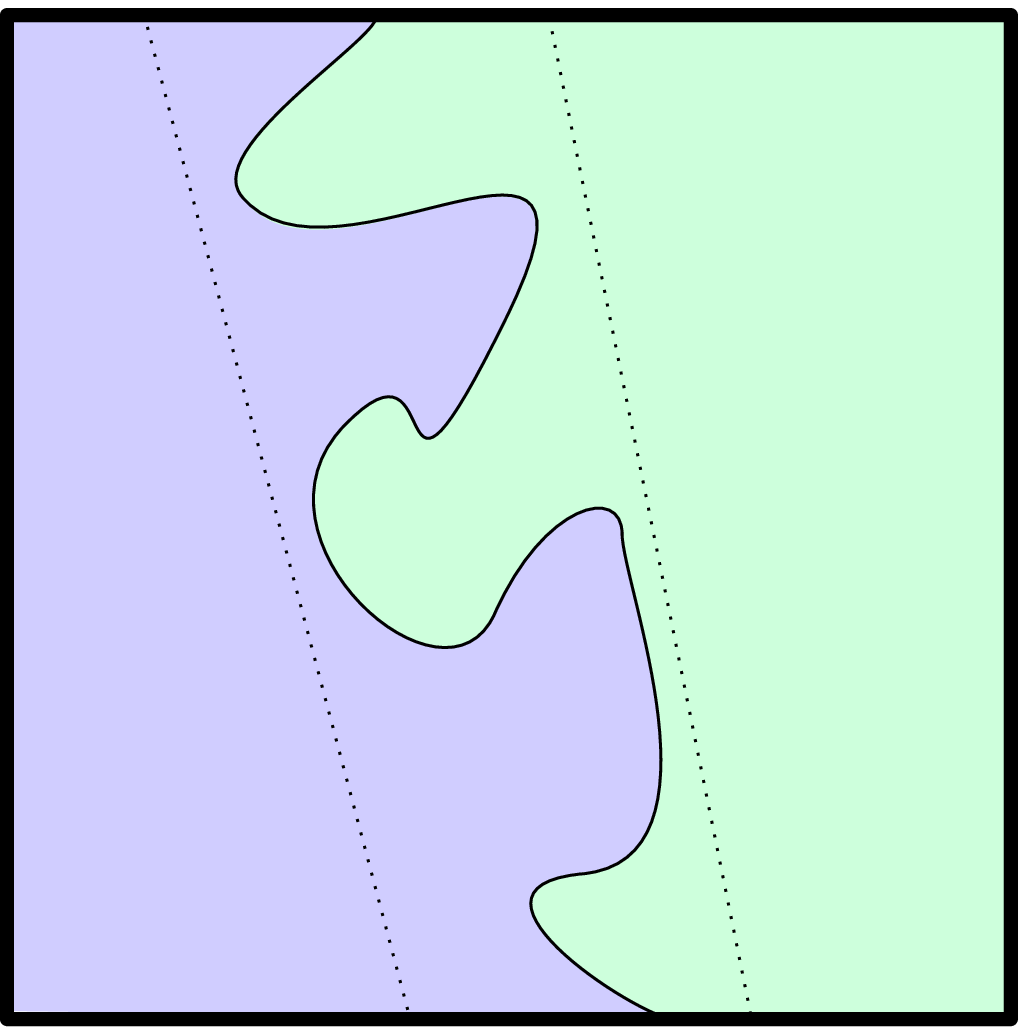}
  \hspace{\stretch{1.3}}
  \caption{A cartoon example of covert overfitting, where the dotted lines
    delimit an ambiguous region of the feature space.  On the left we see the
    ``optimal'' solution and on the right we see a solution where the classifier
    has overfit.  Unlike the case of ordinary overfitting we expect both class
    boundaries to behave similarly in generalization, since in the ambiguous
    region only the volume of the feature space assigned to each class will
    affect performance.}
  \label{fig:covert-overfitting}
\end{figure}

Since it is difficult to make a principled choice of where to place the boundary
in an ambiguous region, the task of identifying covert overfitting is more
difficult than its ordinary counterpart.  Techniques like cross validation,
which estimate the generalization performance by testing the classifier on data
which was not used during training, are able to detect overfitting in
\emph{un}ambiguous regions since an overfit model will not generalize to good
performance on the test data.  Contrastingly, in ambiguous regions, many
different boundaries will achieve comparable generalization performance, since
the posterior class probabilities in these regions are nearly equal.  This means
that we cannot distinguish between parsimonious and overfit solutions in
ambiguous regions based on generalization performance alone.

We demonstrate that covert overfitting occurs using two different methods.  Both
of these methods rely on our ability to apply different degrees of smoothing to
the boundary produced by a trained SVM.  We present a regularization technique
here adapted from \cite{Liang2010a} (with previous work appearing in
\cite{Downs2002} and \cite{Liang2008}).  The key insight allowing this method to
function is a result of Liang's work; however, we have enhanced the algorithm to
allow SVM approximations using an arbitrary number of support vectors to be
constructed in a single step.  While the algorithm in~\cite{Liang2010a} removes
one support vector per iteration, we are able to identify a subset of arbitrary
size to remove while still maintaining the important properties of the
algorithm.


\subsection{Spectral Reduction}
\label{sec:spec-reduction}

Given an SVM, we can express the hyperplane normal vector $\vec{w}$, as a
function of the support vectors \citep[chap.\ 7.3]{Scholkopf2002}.  Let the
support vectors be indexed by a set $N$ and suppose we can partition $N$ into
two disjoint subsets, $I$ and $D$, such that $\mathcal{I} = \{ \vec{x}_i : i \in
I\}$ is a linearly independent set and the elements of $\mathcal{D} =
\{\vec{x}_j : j \in D\}$ are linearly dependent on the elements of
$\mathcal{I}$.  Also, define the function
$\operatorname{Proj}_{\mathcal{I}}(\vec{x})$ as the projection of $\vec{x}$ into
the span of $\mathcal{I}$.  Following\footnote{A very similar derivation for the
  removal of a single support vector appears in \cite{Liang2010a}.  The
  derivation here has been rephrased in terms of the hyperplane normal vector,
  and slightly generalized to account for the removal of several support vectors
  at once.} Liang, we can write:
\begin{align*}
  \vec{w} &= \sum_{i\in N} \alpha_iy_i\vec{x}_i \\
  &= \sum_{i\in I} \alpha_iy_i\vec{x}_i + \sum_{j\in D} \alpha_jy_j\vec{x}_j \\
  &= \sum_{i\in I} \alpha_iy_i\vec{x}_i + \sum_{j\in D}
  \alpha_jy_j\operatorname{Proj}_{\mathcal{I}}(\vec{x}_j) \\
  &= \sum_{i\in I} \alpha_iy_i\vec{x}_i + \sum_{j\in
    D}\alpha_jy_j\bigg(\sum_{i\in I} \beta_{ji}\vec{x}_i\bigg) \\
  &= \sum_{i\in I} \alpha_iy_i\vec{x}_i + \sum_{i\in I}\sum_{j\in
    D}\alpha_jy_j\beta_{ji}\vec{x}_i \\
  &= \sum_{i\in I}\bigg( \alpha_iy_i + \sum_{j\in D}\alpha_jy_j\beta_{ji}
  \bigg)\vec{x}_i \\
  &\triangleq \sum_{i \in I}(\alpha_iy_i)'\vec{x}_i \enspace,
\end{align*}
where the last equality defines $(\alpha_iy_i)'$.  Here $\beta_{ji}$ represents
the $i$th coordinate of $\vec{x}_j$ with respect to $\mathcal{I}$.  This
derivation shows that any linearly dependent support vectors can be eliminated
from the SVM by making an appropriate change to the Lagrange multipliers for the
remaining independent support vectors.  If we restrict $|\mathcal{I}|$ in the
above derivation to be less than the dimensionality of the span of the support
vectors then the third equality becomes an approximation (since $\mathcal{D}$
will no longer be linearly dependent on $\mathcal{I}$) and we find, following
Liang, that provided we select $\mathcal{I}$ so as to minimize $\sum_{j\in D}
||\operatorname{Proj}_{\mathcal{I}}(\vec{x}_j) - \vec{x}_j||$, the resulting SVM
is the best approximation of the original, using $|\mathcal{I}|$ support
vectors.

It is important to note at this point that the $\vec{x}_i$ in the above
derivation must be expressed in the implicit space induced by the kernel.  This
complicates matters since this space may be very high, or even infinite,
dimensional.  Thus, we need a method which does not require us to compute
explicit representations for the support vectors in the implicit space.

The solution to this problem is offered by the kernel matrix.  The kernel matrix
for an SVM with $n$ support vectors is an $n\times n$ symmetric matrix
$\vec{Q}$, such that
\begin{align*}
  \vec{Q}_{ij} = K(\vec{x}_i,\vec{x}_j) \enspace,
\end{align*}
where the $\vec{x}_i$ are the support vectors and $K(\cdot,\cdot)$ is the kernel
function.  The kernel matrix is the Gram matrix of the support vectors, after
applying the implicit mapping implied by the kernel function, and encodes
important information about the SVM.  For instance, since the kernel matrix is a
Gram matrix, $\rank(\vec{Q})$ is equal to the number of linearly independent
support vectors.  Furthermore, if we find a linearly independent spanning subset
of the rows of $\vec{Q}$, we can take the corresponding support vectors as a
minimal set of support vectors required to re-express $\vec{w}$ as in the above
derivation.  In this way the original problem is reduced to finding a subset of
the rows of $\vec{Q}$ which form a basis for its row space.

This basis can be found efficiently by computing the $\vec{LUP}$ decomposition
of $\vec{Q}$.  This gives a lower triangular matrix $\vec{L}$, an upper
triangular matrix $\vec{U}$, and a permutation matrix $\vec{P}$, such that
$\vec{P}\vec{Q} = \vec{L}\vec{U}$.  The matrices $\vec{L}$ and $\vec{U}$ are not
useful to us; however, the matrix $\vec{P}\vec{Q}$ has the useful property that
its first $\rank(\vec{Q})$ rows are linearly independent.  Since $\vec{P}$ is a
permutation matrix, we see immediately that we can use it to identify the rows
of $\vec{Q}$ we require.

The preceding paragraph shows that we can use a linearly independent subset of
the rows of $\vec{Q}$ to select a minimal set of support vectors which can be
used to produce an exact reconstruction of the original SVM.  We now address the
problem of identifying which support vectors we can remove to produce an optimal
rank-reduced approximation of the original.  The goal is to be able to select an
arbitrary number of support vectors and to have a method which we can use to
construct the best possible approximation to our original SVM using the
specified number of support vectors, selected from among the support vectors of
the original.

The key here is to notice that, since $\vec{Q}$ is a symmetric matrix, we can
take its eigenvalue decomposition
\begin{align*}
  \vec{Q} = \vec{V}\vec{\Lambda}\vec{V}^\T \enspace,
\end{align*}
where $\vec{V} = [ \vec{v}_1 \cdots \vec{v}_n ]$ is an orthogonal matrix of
eigenvectors and $\vec{\Lambda} = \diag(\lambda_1, \cdots, \lambda_n)$ is a
diagonal matrix of eigenvalues.  For convenience we can require the eigenvalues
are ordered such that $\lambda_i \ge \lambda_{i+1}$.  If we let $r =
\rank(\vec{Q})$ then we can rewrite this decomposition as
\begin{align}
  \vec{Q} = \sum_{i=1}^n\lambda_i\vec{v}_i\vec{v}_i^\T
  = \sum_{i=1}^r\lambda_i\vec{v}_i\vec{v}_i^\T \enspace,
  \label{eq:eigensum}
\end{align}
where the second equality holds since $\lambda_{r+1} = \dotsi = \lambda_n = 0$.
We can use (\ref{eq:eigensum}) to form approximations of $\vec{Q}$ by truncating
the sum after some $r' < r$ terms, giving
\begin{align*}
  \vec{Q}'= \sum_{i=1}^{r'}\lambda_i\vec{v}_i\vec{v}_i^\T \enspace,
\end{align*}
which is the best rank-$r'$ approximation of $\vec{Q}$.

Since $\vec{Q}'$ is an $n\times n$ matrix with rank $r' < n$ we can select $r'$
linearly independent rows of $\vec{Q}'$ which give a basis for its row space.
Since $\vec{Q}'$ is the best rank-$r'$ approximation of $\vec{Q}$, it follows
that the $r-r'$ dimensions of $\vec{Q}$'s row space not represented in
$\vec{Q}'$ are the dimensions which provide the least contribution to $\vec{Q}$.
Since there is a 1-1 correspondence between the dimensionality of the kernel row
space and the number of support vectors required to represent the SVM
hyperplane, selecting linearly independent rows of $\vec{Q}'$ corresponds to
selecting support vectors whose presence has a large effect on the hyperplane.

We now have sufficient information to construct rank-reduced approximations of a
given SVM.  Training an SVM in the usual way gives us a set of support vectors
and their corresponding Lagrange multipliers.  To construct an approximation of
this SVM using $r'$ support vectors we construct the kernel matrix, $\vec{Q}$
and its best rank-$r'$ approximation, $\vec{Q}'$.  Identifying a subset of the
rows of $\vec{Q}'$ which form a basis for its row space tells us which of the
support vectors to keep in the reduced model (there will be exactly
$r'=\rank(\vec{Q}')$ of them).  We then update the Lagrange multipliers using
the rule,
\begin{align}
  (\alpha_iy_i)' \coloneqq \alpha_iy_i + \sum_{j\in D}\alpha_jy_j\beta_{ji}
  \enspace.
  \label{eq:lagrange}
\end{align}
The new SVM, with support vectors selected using the $\vec{LUP}$ decomposition
of $\vec{Q}'$ and Lagrange multipliers given by (\ref{eq:lagrange}), is the best
approximation of the original SVM using $r'$ support vectors.

The procedure described in this section can be used to produce arbitrary
rank-reduced approximations of a trained SVM.  This gives us access to an entire
spectrum of increasingly more regularized versions of the SVM model.  In the
following sections we exploit this ability to gradually regularize our model in
order to demonstrate the existence of covert overfitting.


\subsection{Hyperplane Angles}

The SVM is, at its core, a linear classifier.  The ability to handle non-linear
problems comes from the kernel, which performs an implicit mapping into a high
dimensional feature space.  In this implicit space, the SVM decision boundary is
represented as the zero level-set of a linear function.  Since the function is
linear, it can be described by its normal vector and so the similarity of two
SVM models can be measured by the angle between the normal vectors of their
corresponding hyperplanes.

We must avoid computing the normal vectors directly, since the dimensionality of
the implicit space may be very high or even infinite.  Nonetheless, it is still
possible to compute the angle between two SVM hyperplanes in the implicit space
without computing their representations explicitly.

In general, ignoring the constant term for simplicity, an SVM hyperplane is
given by
\begin{align*}
  f(\vec{x}) = \sum_{i=1}^r \alpha_iy_i\langle \vec{x}_i,\vec{x} \rangle =
  \vec{\alpha}\vec{X}\vec{x}^\T = \vec{w}\vec{x}^\T
  \enspace,
\end{align*}
where $\vec{\alpha}_i = \alpha_iy_i$, $\vec{X}$ is a matrix with the support
vectors (represented in the implicit space) as its rows, and $\vec{w}$ is the
hyperplane normal vector.  Crucially, the final equality shows that $\vec{w} =
\vec{\alpha}\vec{X}$, which we do not want to compute directly (since $\vec{X}$
is a matrix of vectors in the implicit space), but we can use to compute the
inner product of hyperplane normals.

Suppose now that we have two SVMs, with hyperplane normals given by $\vec{w}_1 =
\vec{\alpha}_1\vec{X}_1$ and $\vec{w}_2 = \vec{\alpha}_2\vec{X}_2$ the angle
between them is
\begin{align*}
  \cos(\theta) &=
  \frac{\vec{w}_1\vec{w}_2^\T}{\vec{w}_1\vec{w}_1^\T\vec{w}_2\vec{w}_2^\T}
  = \frac{\vec{\alpha}_1\vec{X}_1\vec{X}_2^\T\vec{\alpha}_2^\T}
  {\vec{\alpha}_1\vec{X}_1\vec{X}_1^\T\vec{\alpha}_1^\T
    \vec{\alpha}_2\vec{X}_2\vec{X}_2^\T\vec{\alpha}_2^\T}
  \enspace.
\end{align*}
This expression is in terms of the inner products of the rows of $\vec{X}_1$ and
$\vec{X}_2$ (\ie inner products of support vectors in the implicit space) which
can be computed efficiently using the kernel function.  The
$\vec{X}_1\vec{X}_2^\T$ term requires that both SVMs use the same kernel in
order for this method to work.  Since different kernels imply implicit mappings
into different spaces, so the notion of an ``angle'' between the hyperplanes
loses meaning when different kernels are used.

The method described here can be used to measure the angle between an SVM and a
rank-reduced approximation of the same model.  We expect that higher rank
approximations will produce hyperplanes which converge to the original (this
follows directly from our regularization method); however, what we are
interested in is the rate of convergence and more importantly, how the angle
compares to performance.  If covert overfitting is present we expect the
performance of the rank-reduced models to converge to the performance of the
original much faster than the angle between their hyperplanes converges to 0.


\subsection{Class Assignment Variation}

If an SVM has placed its boundary in an ambiguous region, it should be possible
to move the boundary within this region without affecting the performance of the
classifier.  This suggests a method for identifying covert overfitting by
watching for a plateau in performance as the kernel rank is reduced.  Since our
smoothing method guarantees that we move the boundary as little as possible at
each iteration, we expect that the first support vectors to be removed are those
which encode information in the most complex regions of the boundary (which we
expect to correspond to those regions where covert overfitting has occurred).
If these details represent true features of the problem (\ie the true class
boundary is in fact complex in this region) then smoothing the SVM solution will
cause a drop in performance; however, if details removed by the smoothing
process are a result of covert overfitting then we expect the performance to
remain approximately constant as they are removed.

If there are data points near the boundary, it is quite likely that small
changes in the boundary position will cause their predicted label to change.
This will happen regardless of whether or not the boundary correctly encodes the
optimal separating line between the classes.  Thus, we can look for the combined
occurrence of two effects as an indication of covert overfitting:
\begin{enumerate}
\item The SVM rank must be substantially reduced before we see a significant
  drop in performance, and
\item There are many test data which have their predicted label change
  frequently as the rank drops.
\end{enumerate}
Neither of these effects in isolation are sufficient to detect covert
overfitting.  If the classes are highly separated then it may be possible to
reduce the rank substantially without affecting performance, as the boundary is
free to move within the large margin; however, in this case we would not see
variation in label assignment.  Conversely, if we see varying label assignments
but performance drops, then we are likely losing important information about the
true class boundary, rather than details from covert overfitting.  If the
effects are present together then the constant performance indicates that the
overall predictive power of the model is maintained, while at the same time the
label assignment changes indicate that the boundary is moving in a region with a
small margin.


\subsection{Results}
\label{sec:co-results}

In order to demonstrate the existance of covert overfitting we built a synthetic
data set with an overlap level of 0.4 and an imbalance level of 0.6, following
the same procedure as for the previous experiments.  We then trained an SVM
classifier on this data set, using the simulated annealing procedure from
\cite{Boardman2006}, with cross validation to select parameter values.  After an
initial pre-processing step to remove redundant support vectors, we construct a
series of rank-reduced approximations using the method described in
Section~\ref{sec:spec-reduction}.  We use each of these rank-reduced SVMs to
classify a test set drawn from the same generative distribution that was used
for training.  For each rank-reduced SVM we measure the angle between its
hyperplane and that of the original SVM, and record which elements of the test
set have their class assignment change as each support vector is removed.

To decide when the original SVM is sufficiently well approximated by a
rank-reduced approximation, we compare the rank-reduced performance to the
original performance.  We consider the rank-reduced SVMs to be accurate
reconstructions of the original if their test performance is greater than or
equal to $p-\delta$, where $p$ is the performance of the original classifier and
$\delta$ is some small threshold.  We call the lowest-rank for which this occurs
the \emph{sufficiency point} and for our tests we chose $\delta=0.001$.  We are
most interested in the behaviour of the reconstructions with rank greater than
the sufficiency point, as these are the ones which we expect to show variation
within the ambiguous region.

\begin{figure}[tbh]
  \centering
  \subfigure[]{
    \includegraphics[width=0.4\linewidth]
    {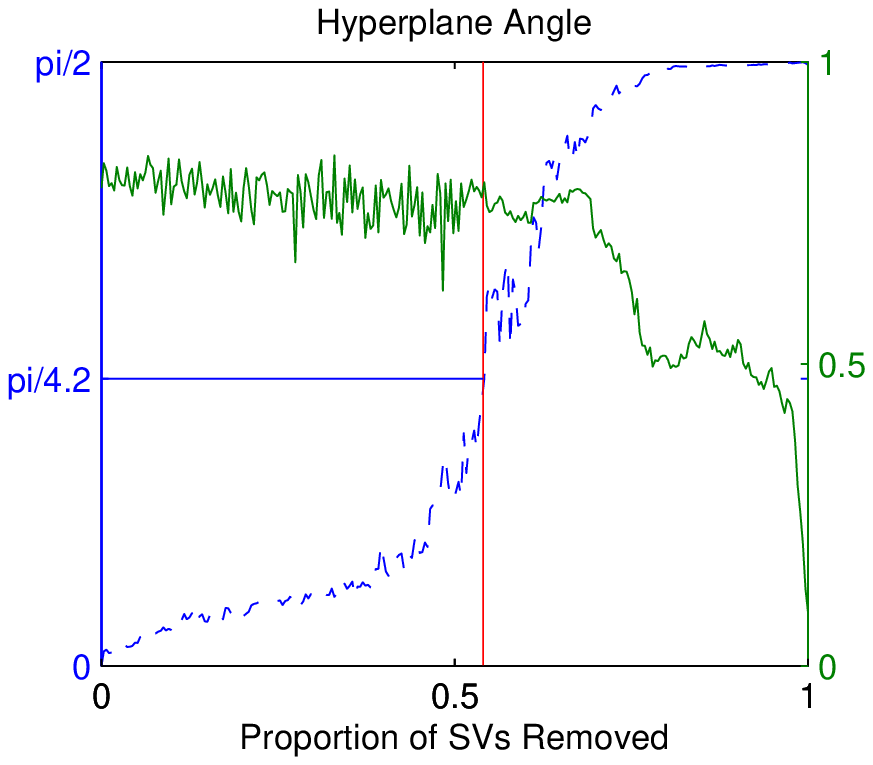}
    \label{fig:angle-v-kern-rank}
  }
  \subfigure[]{
    \includegraphics[width=0.4\linewidth]
    {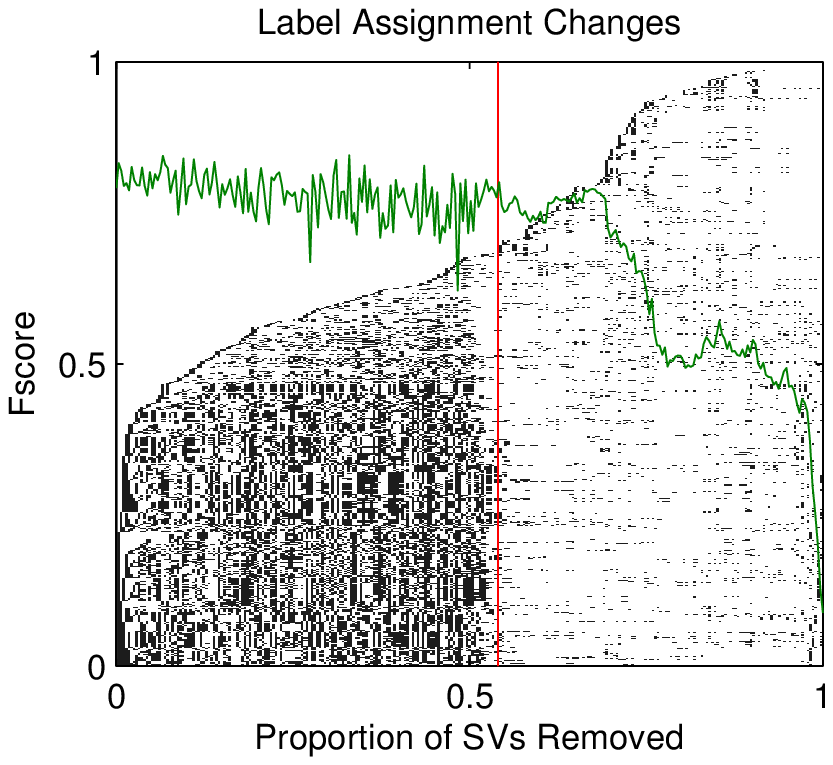}
    \label{fig:label-change-matrix}
  }
  \caption{Covert Overfitting Results. (a) Compares performance (solid line) to
    the angle between the original SVM and it's rank reduced approximations
    (dashed line).  (b) Compares performance to label assignment changes over
    the same domain (see the text for a complete description of this figure).
    The vertical line in both figures indicates the sufficiency point.}
  \label{fig:covert-overfitting-demo}
\end{figure}

Figure~\ref{fig:angle-v-kern-rank} shows an overlaid plot of the performance of
the rank-reduced reconstructions and the angle between the original and
approximated hyperplanes.  The vertical line in the figures shows the
sufficiency point.  What should be immediately striking here is that not only
can more than half the support vectors be removed without significantly altering
the performance, but the angle between the original hyperplane and the
rank-reduced hyperplane at the sufficiency point is quite large.

As the kernel rank increases, the convergence (in angle) of the reconstructed
hyperplanes towards the original is mostly smooth and monotonic, which is
exactly what we expect from the reduction method.  However, since the
performance beyond the sufficiency point is fairly constant, and the angle
between the reconstructed hyperplane and the original at the sufficiency point
is large, it follows that there is a significant amount of information
represented by the original SVM which is not necessary to achieve comparable
performance.

This effect---the representation of additional information beyond what is
required to achieve good performance---is an example of what we expect from
ordinary overfitting.  The difference here is that the test performance is not
reduced by this behaviour, as the ``extra'' information in the training set
which caused the overfitting is present in the test set as well.  Because the
training and test sets exhibit the same systematic problem, we cannot detect
this phenomenon through validation of the performance alone.

Figure~\ref{fig:label-change-matrix} shows the performance of the rank-reduced
SVM approximations overlaid on a visualization of the class assignment variation
as the rank of the reconstruction is changed.  To create this visualization, we
divide the area of the figure into a grid of cells, where the rows correspond to
elements of the test set and the columns correspond to the different kernel
ranks.  Each cell is shaded black if reducing the SVM rank by one causes the
label assigned to the corresponding element of the training set to change.  Note
that this does not indicate if the label is correctly assigned, but instead
tracks when removing a support vector causes the SVM to ``change its mind''
about which label should be assigned to each test instance.  For ease of
interpretation the data have been sorted along the vertical axis, ordered by the
largest rank which causes their label to change.  Again, we are interested in
the behaviour of class assignments when the rank is greater than the sufficiency
point.

In this case we can again see the effects of covert overfitting.  In fact, we
see that the majority of the variation in label assignment takes place after the
sufficiency point, where performance is relatively constant.  We repeated this
experiment on a variety of different backbone models, with varying levels of
overlap and imbalance, and we found that this behaviour is consistent.  The
number of test data whose label is changed before the sufficiency point is high
when there is strong overlap, and the frequency of label assignment changes is
typically densest in this region as well.

What remains unclear at this point is to what degree the variation is localized
to the ambiguous regions.  We have demonstrated that there is movement in the
SVM hyperplane beyond the sufficiency point, and that this hyperplane movement
causes significant changes in how the SVM assigns labels to test data, despite
the performance remaining constant.  However, it is possible that the label
changes we are seeing are spread uniformly across the entire domain.

\begin{figure}[tbh]
  \centering
  \subfigure {
    \includegraphics[width=0.4\linewidth]
    {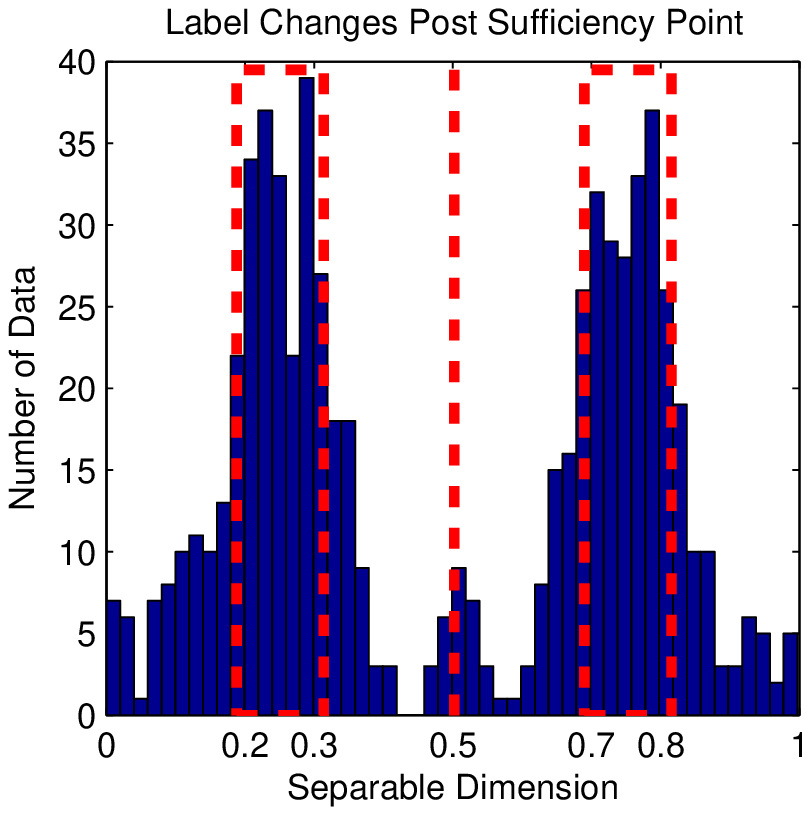}
    \label{fig:label-change-histogram}
  }
  \subfigure {
    \includegraphics[width=0.4\linewidth]
    {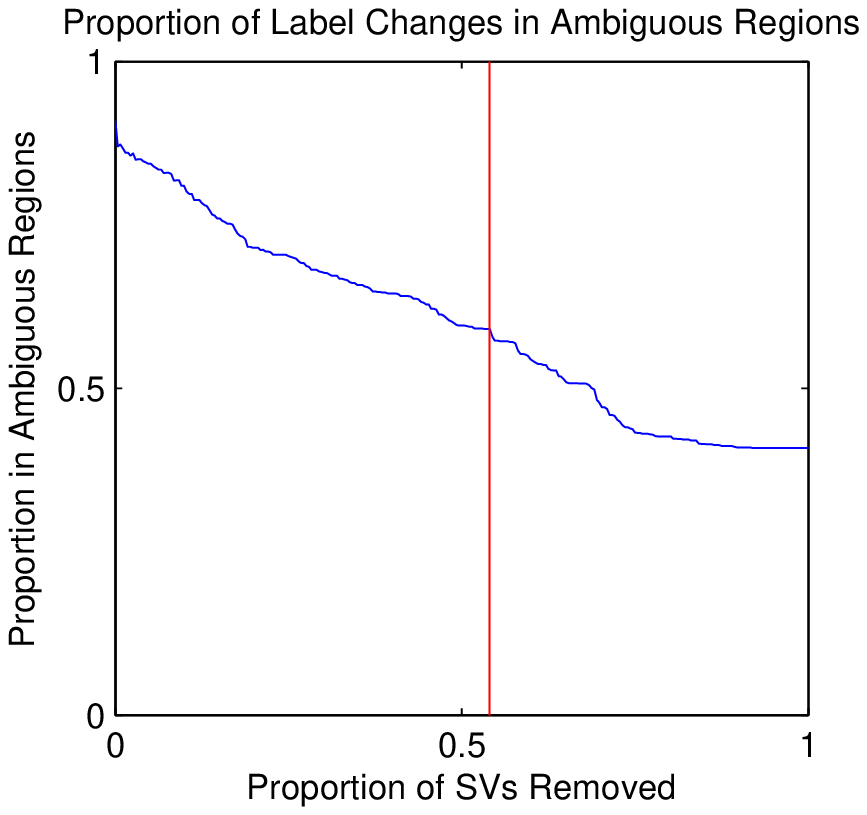}
    \label{fig:label-prop-ambig}
  }
  \caption{A demonstration that covert overfitting is localized to the ambiguous
    regions of the data space. (a) Shows the distribution of test data with at
    least one label change with rank higher than the sufficiency point.  The
    boxes in the diagram delimit ambiguous regions.  (b) Shows the proportion of
    label changes localized to the ambiguous regions at various degrees of
    smoothing.}
  \label{fig:covert-overfitting-variations}
\end{figure}

To show that the label changes are in fact localized in the ambiguous regions,
we select the test data whose label is changed at least once after the
sufficiency point has been reached and check if they are localized to the
ambiguous region.  Figure~\ref{fig:label-change-histogram} shows the
distribution of these data along the dimension in which they are distinguishable
(recall from Section~\ref{sec:data-sets} that our 2D backbone models are
indistinguishable in only one dimension).  The distribution is clearly localized
in the ambiguous regions with some additional variation near the boundaries (\eg
note the behaviour around the crisp boundary at 0.5).

Figure~\ref{fig:label-prop-ambig} demonstrates that the degree of localization
of label variations to the ambiguous regions across several degrees of
smoothing.  The trend line in this figure shows, for each level of smoothing,
the proportion of test data which have had their label assignment change at
least once and lie in an ambiguous region. When the rank is extremely low the
proportion is approximately 0.42, which is equal to the proportion of the entire
test set which lies in an ambiguous region; however, we see that when we
consider high rank approximations the label changes are highly localized to the
ambiguous regions.



\section{Conclusion}

In this paper we first looked at how the overlap and imbalance problems in
isolation affect performance of the SVM classifier.  In the case of imbalance we
saw that when there are sufficiently many training data, imbalance does not
degrade the SVM performance.  We also saw, in the case of overlap in isolation,
that even when there are ambiguous regions in the data space, the SVM is still
able to achieve approximately optimal performance.  Naturally, in this case the
overall performance is significantly lower than the imbalanced case, but this is
a result of inherent ambiguity in the data themselves.  Our experiments show
that despite this ambiguity, the SVM is capable of learning models with
performance comparable to an optimal classifier for these domains.

Although the performance on overlapping domains is quite good (compared to an
optimal classifier), the complexity of the learned models is very high.
Increasing either the size of the training set, or the degree of overlap, in
these cases causes the SVM to learn more complex models.  The increased
complexity indicates a systematic weakness of the SVM classifier in the presence
of overlapping data, since the optimal solution on our overlapped domains has
the same complexity as the separable cases.

We used our performance measurements in the cases of imbalance and overlap in
isolation to predict performance for the combined case, under the assumption
that the factors act independently.  We established, following our previous work
in \cite{Denil2010b}, that there is an interdependency between the effects from
these two factors.

The later sections of this work offer a causal explanation for the behaviour in
performance and complexity that we seen in the case of overlapped, as well as
overlapped and imbalanced data.  Our explanation postulates that the behaviour
we see in these cases is caused by covert overfitting.  In order to test this
explanation we developed an SVM pruning method which allows us to build
arbitrary rank approximations of a given SVM. We described two methods for
exploiting this technique to identify the occurrence of covert overfitting;
first by examining the hyperplane angle between an SVM and its low rank
approximations and second by looking at the frequency and localization of label
assignment changes with respect to the rank of the approximation.  In both cases
our findings are consistent with the occurrence of covert overfitting and
provide evidence that it is a real problem for training high quality SVMs.

We established that when overlapping classes are present in the data a
significant amount of the support vectors in a trained SVM model go towards
encoding aspects of the boundary which do not increase the generalization
performance.  We also saw that the removal of these support vectors produces
variation in class label assignment which is localized around the ambiguous
regions of the data space.  The degree of this localization is highest when the
approximations are near to the original SVM.

One of the original goals of this work was to formulate a measure of overlap in
real world data.  To that end we have identified several characteristics,
notably the relationship between overlap and imbalance, which such a measure
must account for.  We have also identified a specific behaviour, namely covert
overfitting, which we have shown to be indicative of overlapping classes.  We
have demonstrated how this behaviour can be detected through two signature
effects: redundancy in the support vectors of the trained model, and the
variation of class assignments under regularization.  Further work will
investigate if these characteristics can be turned into an overlap measure which
is applicable to real world data.




\vskip 0.2in
\bibliographystyle{plainnat}
\bibliography{biblio}

\begin{thebibliography}{17}
\providecommand{\natexlab}[1]{#1}
\providecommand{\url}[1]{\texttt{#1}}
\expandafter\ifx\csname urlstyle\endcsname\relax
  \providecommand{\doi}[1]{doi: #1}\else
  \providecommand{\doi}{doi: \begingroup \urlstyle{rm}\Url}\fi

\bibitem[Akbani et~al.(2004)Akbani, Kwek, and Japkowicz]{Akbani2004}
Rehan Akbani, Stephen Kwek, and Nathalie Japkowicz.
\newblock {Applying Support Vector Machines to Imbalanced Datasets}.
\newblock \emph{Machine Learning: ECML 2004}, pages 39--50, 2004.

\bibitem[Auda and Kamel(1997)]{Auda1997}
Gasser Auda and Mohamed Kamel.
\newblock {CMNN: Cooperative modular neural networks for pattern recognition}.
\newblock \emph{Pattern Recognition Letters}, 18\penalty0 (11-13):\penalty0
  1391--1398, 1997.

\bibitem[Batista et~al.(2004)Batista, Prati, and Monard]{Monard}
Gustavo Batista, Ronaldo~C Prati, and Maria~Carolina Monard.
\newblock {A Study of the Behavior of Several Methods for Balancing Machine
  Learning Training Data}.
\newblock \emph{SIGKDD Explorations}, 6\penalty0 (1):\penalty0 20--29, 2004.

\bibitem[Batista et~al.(2005)Batista, Prati, and Monard]{Batista2005}
Gustavo Batista, Ronaldo~C Prati, and M.C. Monard.
\newblock {Balancing strategies and class overlapping}.
\newblock \emph{Lecture notes in computer science}, 3646:\penalty0 24, 2005.

\bibitem[Boardman and Trappenberg(2006)]{Boardman2006}
Matthew Boardman and Thomas Trappenberg.
\newblock \emph{{A Heuristic for Free Parameter Optimization with Support
  Vector Machines}}.
\newblock IEEE, 2006.
\newblock ISBN 0-7803-9490-9.
\newblock \doi{10.1109/IJCNN.2006.1716150}.

\bibitem[Bosch et~al.(1997)Bosch, Weijters, Herik, and Daelemans]{Weijters1997}
Antal Van~Den Bosch, Ton Weijters, H.~Jaap Van~Den Herik, and Walter Daelemans.
\newblock {When small disjuncts abound, try lazy learning: A case study}.
\newblock \emph{Proceedings seventh BENELEARN conference}, pages 109--118,
  1997.

\bibitem[Denil and Trappenberg(2010)]{Denil2010b}
Misha Denil and Thomas Trappenberg.
\newblock {Overlap versus Imbalance}.
\newblock In Atefeh Farzindar and Vlado Keselj, editors, \emph{Advances In
  Artificial Intelligence}, volume 6085, pages 220--231, Ottawa, 2010.
  Springer.

\bibitem[Downs et~al.(2002)Downs, Gates, and Masters]{Downs2002}
Tom Downs, Kevin~E. Gates, and Annette Masters.
\newblock {Exact simplification of support vector solutions}.
\newblock \emph{The Journal of Machine Learning Research}, 2:\penalty0 297,
  2002.

\bibitem[Japkowicz(2003)]{Japkowicz2003}
Nathalie Japkowicz.
\newblock {Class imbalances: are we focusing on the right issue}.
\newblock \emph{Workshop on Learning from Imbalanced Data Sets II}, pages
  17--23, 2003.

\bibitem[Japkowicz and Stephen(2002)]{Japkowicz2002}
Nathalie Japkowicz and Shaju Stephen.
\newblock {The class imbalance problem: A systematic study}.
\newblock \emph{Intelligent Data Analysis}, 6:\penalty0 429--449, 2002.

\bibitem[Jo and Japkowicz(2004)]{Jo2004}
Taeho Jo and Nathalie Japkowicz.
\newblock {Class imbalances versus small disjuncts}.
\newblock \emph{ACM SIGKDD Explorations Newsletter}, 6\penalty0 (1):\penalty0
  40--49, 2004.

\bibitem[Liang(2010)]{Liang2010a}
Xun Liang.
\newblock An effective method of pruning support vector machine classifiers.
\newblock \emph{IEEE Transactions on Neural Networks}, 21\penalty0
  (1):\penalty0 26--38, 2010.

\bibitem[Liang et~al.(2008)Liang, Chen, and Guo]{Liang2008}
Xun Liang, R.C. Chen, and Xinyu Guo.
\newblock {Pruning Support Vector Machines Without Altering Performances}.
\newblock \emph{IEEE Transactions on Neural Networks}, 19\penalty0
  (10):\penalty0 1792--1803, 2008.

\bibitem[Prati et~al.(2004)Prati, Batista, and Monard]{Prati2004}
Ronaldo~C Prati, Gustavo Batista, and Maria~C Monard.
\newblock {Class Imbalances versus Class Overlapping: An Analysis of Learning
  System Behavior}.
\newblock \emph{MICAI 2004: Advances in Artificial Intelligence}, pages
  312--321, 2004.

\bibitem[Scholkopf and Smola(2002)]{Scholkopf2002}
Bernhard Scholkopf and Alexander~J. Smola.
\newblock \emph{{Learning With Kernels}}.
\newblock Massachusetts Institute of Technology, 2002.

\bibitem[Visa and Ralescu(2003)]{Visa}
Sofia Visa and Anca Ralescu.
\newblock {Learning imbalanced and overlapping classes using fuzzy sets}.
\newblock In \emph{ICML-2003 Workshop on Learning from Imbalanced Data Sets
  II}, volume~3, 2003.

\bibitem[Yaohua and Jinghuai(2007)]{Yaohua2007}
Tang Yaohua and Gao Jinghuai.
\newblock {Improved Classification for Problem Involving Overlapping Patterns}.
\newblock \emph{IEICE Transactions on Information and Systems}, 90\penalty0
  (11):\penalty0 1787--1795, 2007.

\end{thebibliography}

\end{document}